\documentclass[10pt,a4paper]{article}
\usepackage{graphicx}
\usepackage{subfigure}
\usepackage{amsmath,amsthm}
\usepackage{amssymb,amsfonts}
\usepackage{authblk}
\usepackage{url}
\usepackage{hyperref}
\usepackage{multirow}
\usepackage{subfigure}
\usepackage{booktabs}

\newtheorem{definition}{\textbf{Definition}}

\newtheorem{theorem}{\textbf{Theorem}}

\newtheorem{corollary}{\textbf{Corollary}}

\title{Facies Classification with Copula Entropy}
\author{Jian MA\thanks{Email: majian03@gmail.com}}
\date{}

\begin{document}
\maketitle

\begin{abstract}
In this paper we propose to apply copula entropy (CE) to facies classification. In our method, the correlations between geological variables and facies classes are measured with CE and then the variables associated with large negative CEs are selected for classification. We verified the proposed method on a typical facies dataset for facies classification and the experimental results show that the proposed method can select less geological variables for facies classification without sacrificing classification performance. The geological variables such selected are also interpretable to geologists with geological meanings due to the rigorous definition of CE.

\end{abstract}

\section{Introduction}
Facies are the type of rocks with similar characteristics given by geologists and facies classification is of very significance in geological tasks, such as formation evaluation, reservoir characterization. As the geological data accumulates, there are growing interests in facies classification with machine learning methods \cite{Bohling2003,Dubois2007,Hall2016,Alaudah2019,Dramsch2018,Zhao2018,Babai2025,Rahimi2022,Gkortsas2019}.

There are two issues with the existing works on facies classification. First, the machine learning models are built without variable selection or with only very primary method, such as cross-validation, which makes the classifiers with useless variable as inputs and therefore with low performance. Second, most of the models for facies classification are block-box, such as deep learning \cite{Dramsch2018,Noh2023,Saikia2019}, Boostings or SVMs\cite{Babai2025}, which are un-interpretable to geologists.

Variable selection is a common task that selects a subset from all the available variables for machine learning models. By this, the accuracy of the predictive models built with the selected variables can be improved compared with those built without selection. The traditional method for variable selection are mainly based on likelihoods, such as AIC, BIC, or accuracy, such as LASSO \cite{Tibshirani1996}, or correlation, such as HSIC \cite{Gretton2007}, distance correlation \cite{Szekely2007}, and copula entropy \cite{Ma2021a}.

Copula Entropy (CE) is a recently proposed rigorous mathematical concept for measuring multivariate statistical independence and is proved to be equivalent to mutual information in information theory \cite{Ma2011}. Compared with traditional correlation measures, such as the Pearson correlation coefficient, which can only be applied to linear cases with Gaussianity assumption, CE can be applied to any cases without any assumption on the underlying distribution of random variables. A non-parametric method for estimating copula entropy has also been proposed based on rank statistic \cite{Ma2011}. CE-based variable selection has been proposed \cite{Ma2021a} and applied in many scientific fields, such as hydrology \cite{Chen2013}, medicine \cite{Mesiar2021}, among others.

In this paper we propose to apply CE to facies classification. The correlations between geological variables and facies classes are measured with CE and then the geological variables are selected according to their corresponding CE values for training classifiers. In this way, CE benefits facies classification on several aspects. First, CE is model-free, which makes it applicable to the nonlinear relationships between geological variables and facies classes; second, CE is an information theoretical measures for statistical independence, which makes the selected variables interpretable to geologists; third, CE is rigorous defined and with mature non-parametric estimator, which makes it reliable for data analysis, not only for facies classification, but also for other similar geological tasks.

We applied our method to a typical dataset \cite{Dubois2007} which has been widely studied by others. The experimental results verified the effectiveness of our method by correctly selecting less geological variables for classifiers without sacrificing the performance. Meanwhile, the selection results are also interpretable.

This paper is organized as follows: Section \ref{s:method} introduces the methodology, Section \ref{s:ours} presents our method for facies classification with CE, Section \ref{s:exp} presents experiments and results, followed by some discussion \ref{sec:discussion}, finally we conclude the paper in Section \ref{sec:con}.

\section{Methodology}
\label{s:method}
\subsection{Copula Entropy}
\label{s:CopEnt}
\subsubsection{Theory}
Copula theory is about the representation of multivariate dependence with copula function \cite{joe2014,nelsen2007}. At the core of copula theory is Sklar's theorem \cite{sklar1959} which states that multivariate probability density function can be represented as a product of its marginals and copula density function which represents dependence structure among random variables. Such representation separates dependence structure, i.e., copula function, with the properties of individual variables -- marginals, which make it possible to deal with dependence structure only regardless of joint distribution and marginal distribution. This section is to define an statistical independence measure with copula. For clarity, please refer to \cite{Ma2011} for notations.

With copula density, Copula Entropy is define as follows \cite{Ma2011}:
\begin{definition}[Copula Entropy]
	\label{d:ce}
	Let $\mathbf{X}$ be random variables with marginal distributions $\mathbf{u}$ and copula density $c(\mathbf{u})$. CE of $\mathbf{X}$ is defined as
	\begin{equation}
	H_c(\mathbf{X})=-\int_{\mathbf{u}}{c(\mathbf{u})\log{c(\mathbf{u})}}d\mathbf{u}.
	\end{equation}
\end{definition}

In information theory, MI and entropy are two different concepts \cite{Cover1999}. In \cite{Ma2011}, Ma and Sun proved that they are essentially same -- MI is also a kind of entropy, negative CE, which is stated as follows: 
\begin{theorem}
	\label{thm1}
	MI of random variables is equivalent to negative CE:
	\begin{equation}
	I(\mathbf{X})=-H_c(\mathbf{X}).
	\end{equation}
\end{theorem}
\noindent
The proof of Theorem \ref{thm1} is simple \cite{Ma2011}. There is also an instant corollary (Corollary \ref{c:ce}) on the relationship between information of joint probability density function, marginal density function and copula density function.
\begin{corollary}
	\label{c:ce}
	\begin{equation}
	H(\mathbf{X})=\sum_{i}{H(X_i)}+H_c(\mathbf{X}).
	\end{equation}
\end{corollary}
The above results cast insight into the relationship between entropy, MI, and copula through CE, and therefore build a bridge between information theory and copula theory. CE itself provides a mathematical theory of statistical independence measure.

\subsubsection{Estimation}
\label{s:est}
It has been widely considered that estimating MI is notoriously difficult. Under the blessing of Theorem \ref{thm1}, Ma and Sun \cite{Ma2011} proposed a simple and elegant non-parametric method for estimating CE (MI) from data which comprises of only two steps\footnote{The \textsf{R} package \textsf{copent} for estimating CE is available on CRAN and also on GitHub at \url{https://github.com/majianthu/copent}.}:
\begin{enumerate}
	\item Estimating Empirical Copula Density (ECD);
	\item Estimating CE.
\end{enumerate}

For Step 1, if given data samples $\{\mathbf{x}_1,\ldots,\mathbf{x}_T\}$ i.i.d. generated from random variables $\mathbf{X}=\{x_1,\ldots,x_N\}^T$, one can easily estimate ECD as follows:
\begin{equation}
F_i(x_i)=\frac{1}{T}\sum_{t=1}^{T}{\chi(\mathbf{x}_{t}^{i}\leq x_i)},
\end{equation}
where $i=1,\ldots,N$ and $\chi$ represents for indicator function. Let $\mathbf{u}=[F_1,\ldots,F_N]$, and then one can derive a new samples set $\{\mathbf{u}_1,\ldots,\mathbf{u}_T\}$ as data from ECD $c(\mathbf{u})$. In practice, Step 1 can be easily implemented non-parametrically with rank statistic.

Once ECD is estimated, Step 2 is essentially a problem of entropy estimation which has been contributed with many existing methods. Among them, the kNN method \cite{Kraskov2004} was suggested in \cite{Ma2011}. With rank statistic and the kNN method, one can derive a non-parametric method of estimating CE, which can be applied to any situation without any assumption on the underlying system.

\subsection{Random Forests}
Random Forests (RFs) is another widely-used machine learning algorithms developed by Leo Breiman \cite{Breiman2001}. It learn a model from data by ensembling a group of decision trees. It enjoys a good ability of generalization and model flexibility compared with other machine learning algorithms. 

\section{Facies classification with CE}
\label{s:ours}
In this work we first use CE to measure the correlations between geological variables and facies classes and then select these variables according to their corresponding negative CE values. The selected variables are then used for training RFs for classification. 

Our method with CE has many merits compared with traditional method. Since CE is model-free, our method is applicable without any assumption on the underlying distribution of data. As a information theoretical measure, CE measures the information carried by geological variable about facies and therefore makes our method interpretable and more advantageous than traditional black-box methods.

\section{Experiments and Results}
\label{s:exp}

\subsection{The Facies Data}
\begin{figure}
	\centering
	\includegraphics[width=0.6\textwidth]{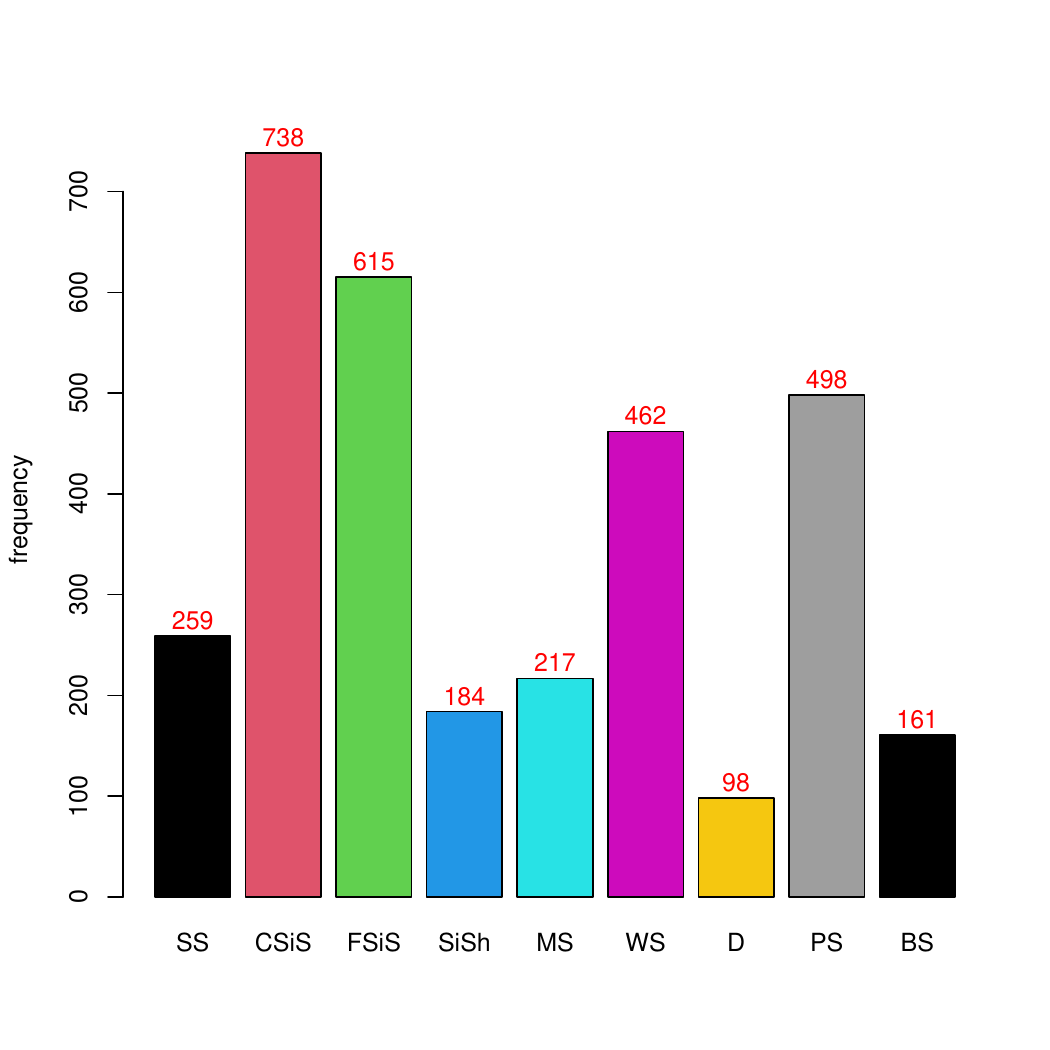}
	\caption{The distribution of the classes of rock facies in the experimental data.}
	\label{fig:classes}
\end{figure}

\begin{figure}
	\centering
	\includegraphics[width=\textwidth]{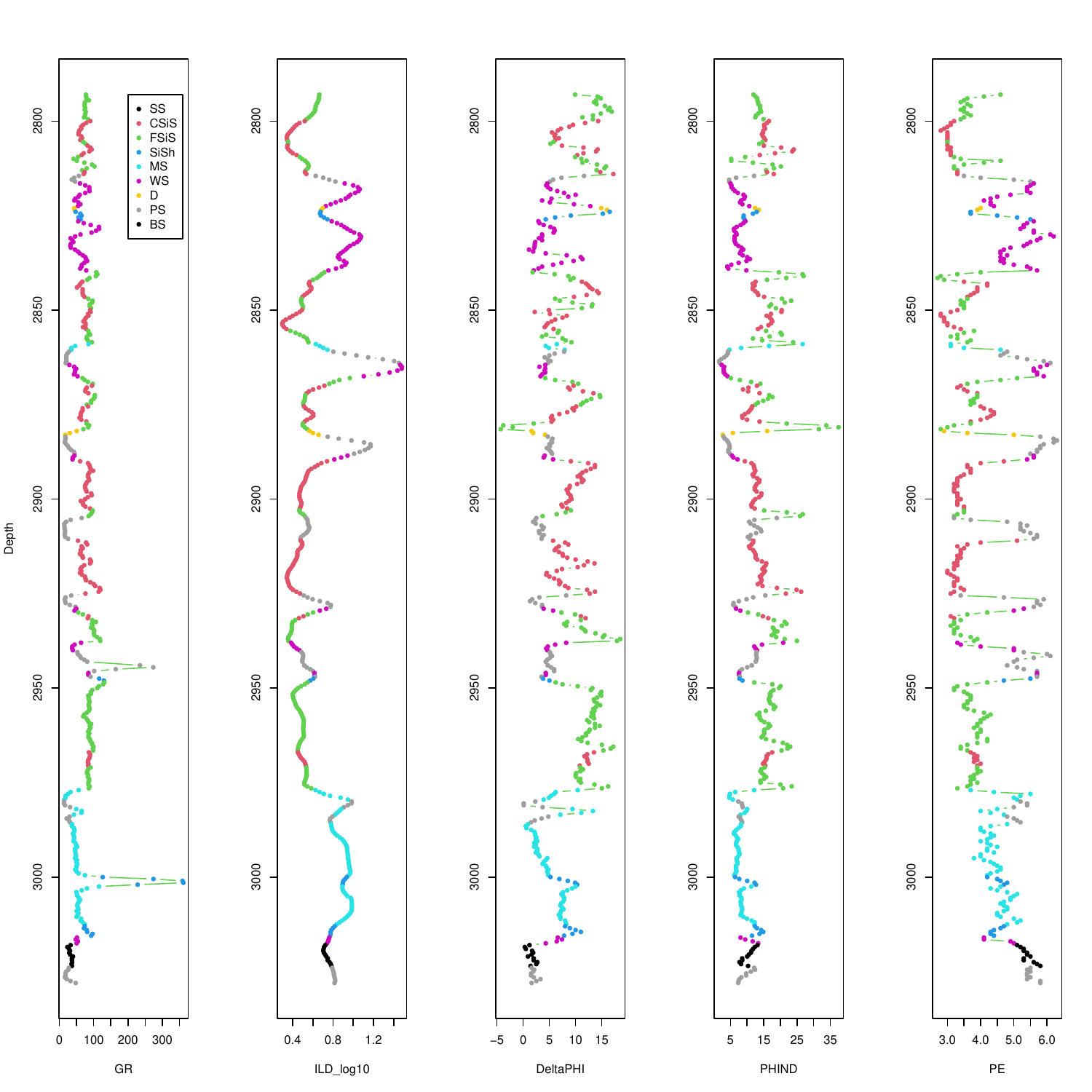}
	\caption{The 5 variable of the well ``SHRIMPLIN".}
	\label{fig:shrimplin}
\end{figure}

\begin{figure}
	\centering
	\includegraphics[width=\textwidth]{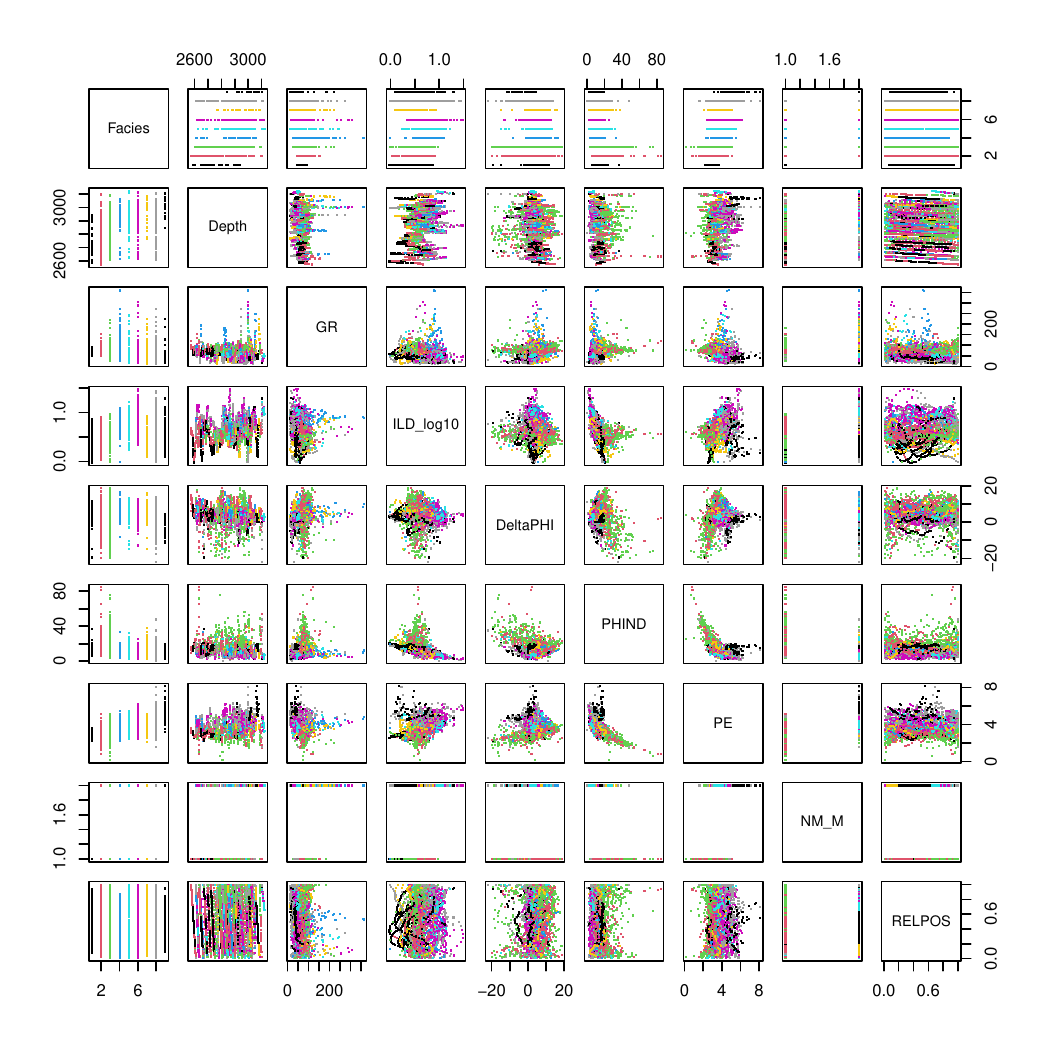}
	\caption{Joint distributions between geological variables and facies classes in the experimental data.}
	\label{fig:pairs}
\end{figure}

The data used in the experiments is obtained from the Council Grove gas reservoir in Southwest Kansas \cite{Bohling2003,Dubois2007}. The dataset used here includes 3232 samples that are derived from 8 wells and each sample consist of 7 variables, of which 5 variables including gamma ray (GR), resistivity logging (ILD\_log10), photoelectric effect (PE), neutron-density porosity difference and average neutron-density porosity (DeltaPHI and PHIND) are wireline log measurements and, 2 variables including nonmarine-marine indicator (NM\_M) and relative position (RELPOS) are geologic constraining variables based on geological knowledge, and also a rock facies labels (with 9 classes as listed in Table \ref{t:classes}) given after investigations on the cores from these wells by half-foot intervals vertically. This distribution of the 9 classes of rock facies is shown in Figure \ref{fig:classes}. Figure \ref{fig:shrimplin} shows the data of 5 variables of a well colored with rock facies classes. This data has been studied for facies classification in \cite{Dubois2007} among others. The joint distributions between the 7 geological variables and facies classes are shown in Figure \ref{fig:pairs}.

\begin{table}
	\centering
	\caption{The rock facies classes of the well logs.}
	\begin{tabular}{l|c}
		\toprule
		Name&Meaning\\
		\midrule
		SS&Nonmarine sandstone\\
		CSiS& Nonmarine coarse siltstone\\
		FSiS&Nonmarine fine siltstone\\
		SiSH&Marine siltstone and shale\\
		MS&Mudstone (limestone)\\
		WS&Wackestone (limestone)\\
		D&Dolomite\\
		PS&Packstone-grainstone (limestone)\\
		BS&Phylloid-algal bafflestone (limestone)\\
		\bottomrule		
	\end{tabular}
	\label{t:classes}
\end{table}

\subsection{Experiments}
In this experiment we apply CE to facies classification by means of selecting inputs of classifiers with it. We first estimated the values of CE between 8 geological variables and facies label with the non-parametric method of CE. The larger the negative CE value, the more important the variable. The variables are ranked according to their corresponding negative CE values and 7 variable sets are selected for classification.

After estimating CEs, we built the facies classifiers with RFs with all the experimental data. The dataset is separated into training and testing data by 80/20 ratios randomly and then a RFs classifier was trained and tested. The default hyper-parameter setting was used in the experiments. Two types of performance measures were used: mean average error (MAE) and prediction accuracy. We derived two types of performance accuracy: the original accuracy which is based on only facies class labels, and the adjusted accuracy which is derived by considering the adjacency of facies classes. In the adjusted accuracy, a prediction is considered also as correct if a prediction of a classifier is an adjacent facies of the target facies. The adjacency relationships of facies according to geological knowledge are shown in Table \ref{t:adj}. The above experiment were conducted 30 times to derive the average performance measures.

The \textsf{R} package \texttt{copent} \cite{Ma2021} and \texttt{randomForest} were used in the experiments as the implementations of the CE estimator and Random Forests respectively.

\begin{table}
	\centering
	\caption{The adjacency relationships of facies according to geological knowledge.}
	\begin{tabular}{l|c}
		\toprule
		Facies&Adjacent Facies\\
		\midrule
		SS&CSiS\\
		CSiS&SS,FSiS\\
		FSiS&CSiS\\
		SiSh&MS\\
		MS&SiSh,WS\\
		WS&MS,D\\
		D&WS,PS\\
		PS&WS,D,BS\\
		BS&D,PS\\
		\bottomrule
	\end{tabular}
	\label{t:adj}
\end{table} 

\subsection{Results}
\begin{figure}
	\centering
	\includegraphics[width=0.8\textwidth]{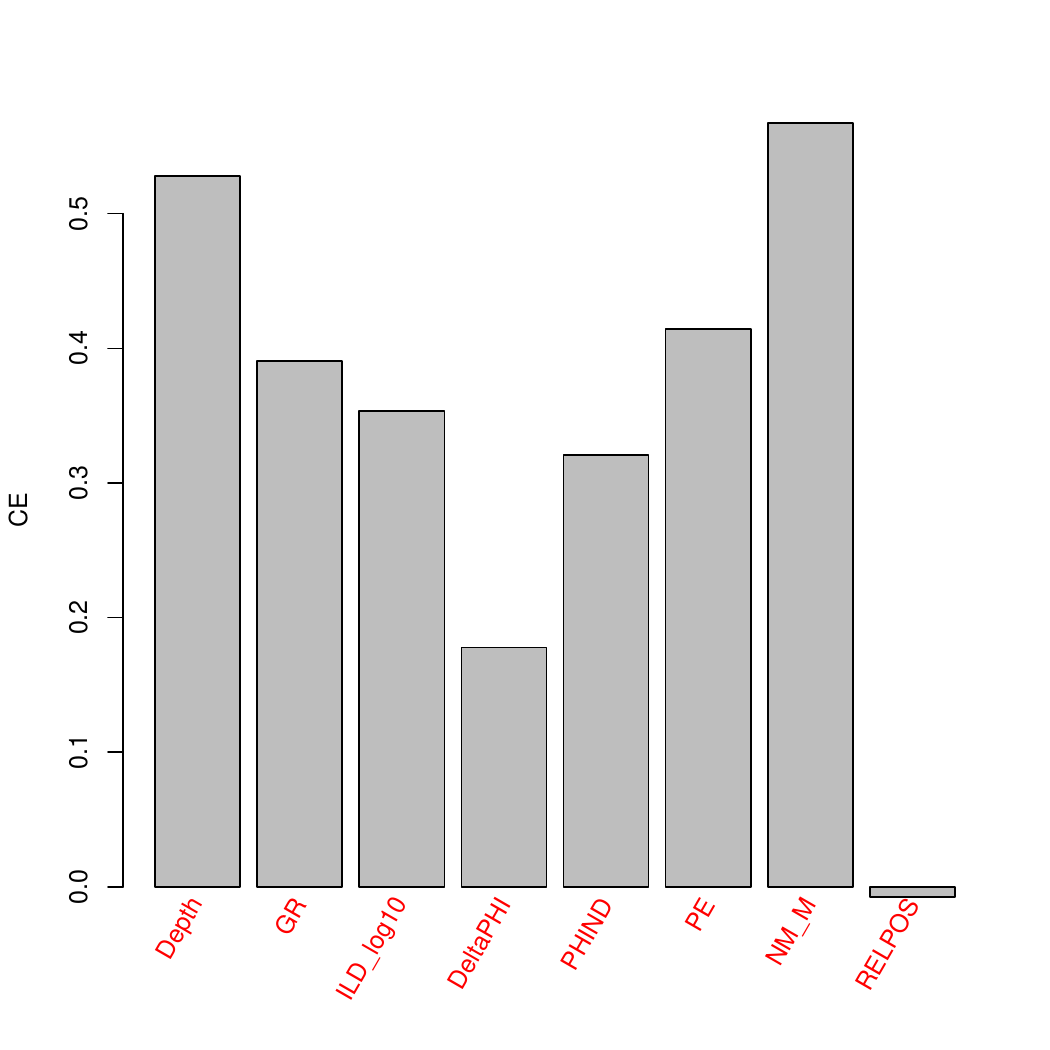}
	\caption{The estimated CEs between geological variables and facies classes.}
	\label{fig:ce}
\end{figure}

The estimated negative CEs between geological variables and facies classes is shown in Figure \ref{fig:ce}. It can be learned from it that the variables ``NM\_M" and ``Depth" present the largest negative CEs while the variables ``RELPOS" and ``DeltaPHI" present the smallest negative CEs.

According to the rank of the estimated negative CEs, we selected 7 groups of variable as inputs of the RFs classifiers, which is shown in Table \ref{t:selectedvars}.

\begin{figure}
	\centering
	\subfigure[MAE]{\includegraphics[width=0.48\textwidth]{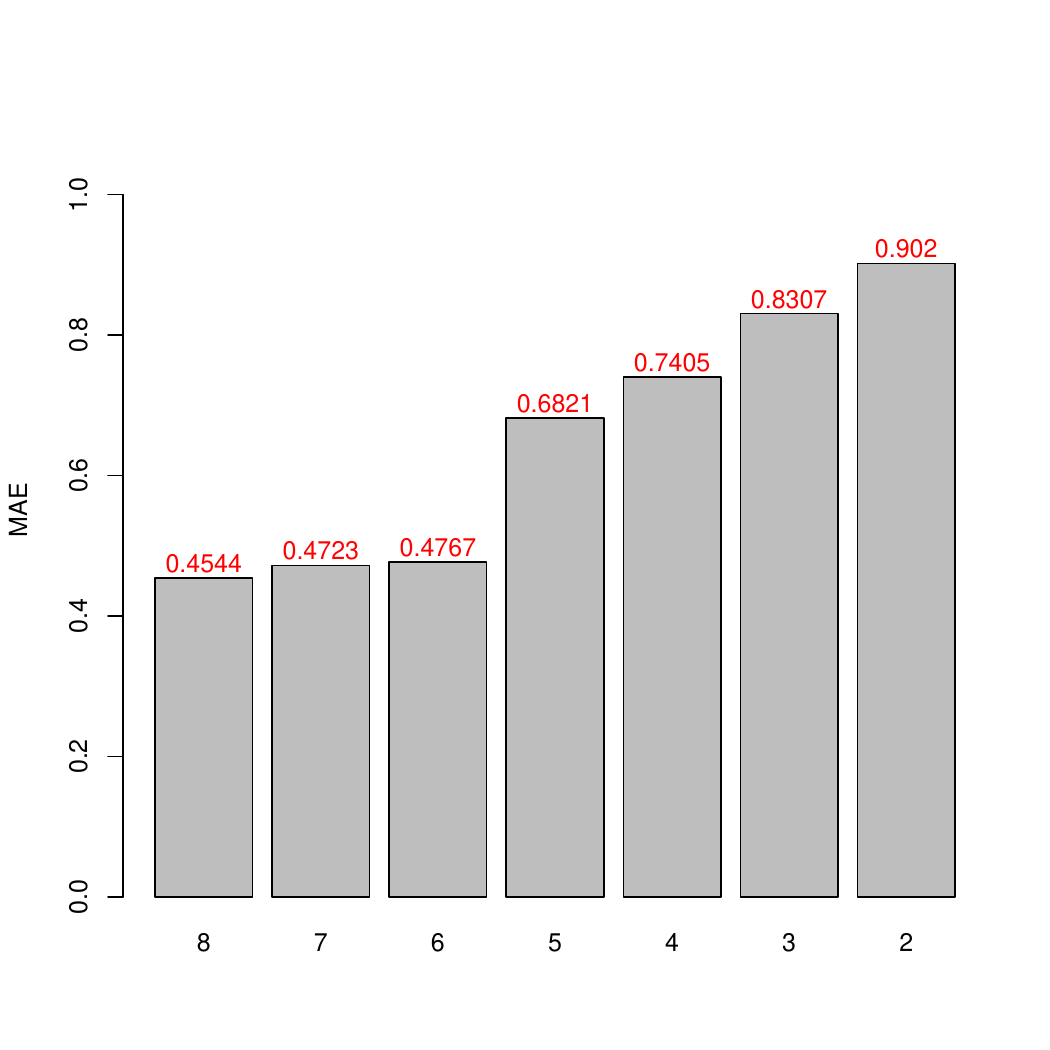}}
	\subfigure[Accuracy]{\includegraphics[width=0.48\textwidth]{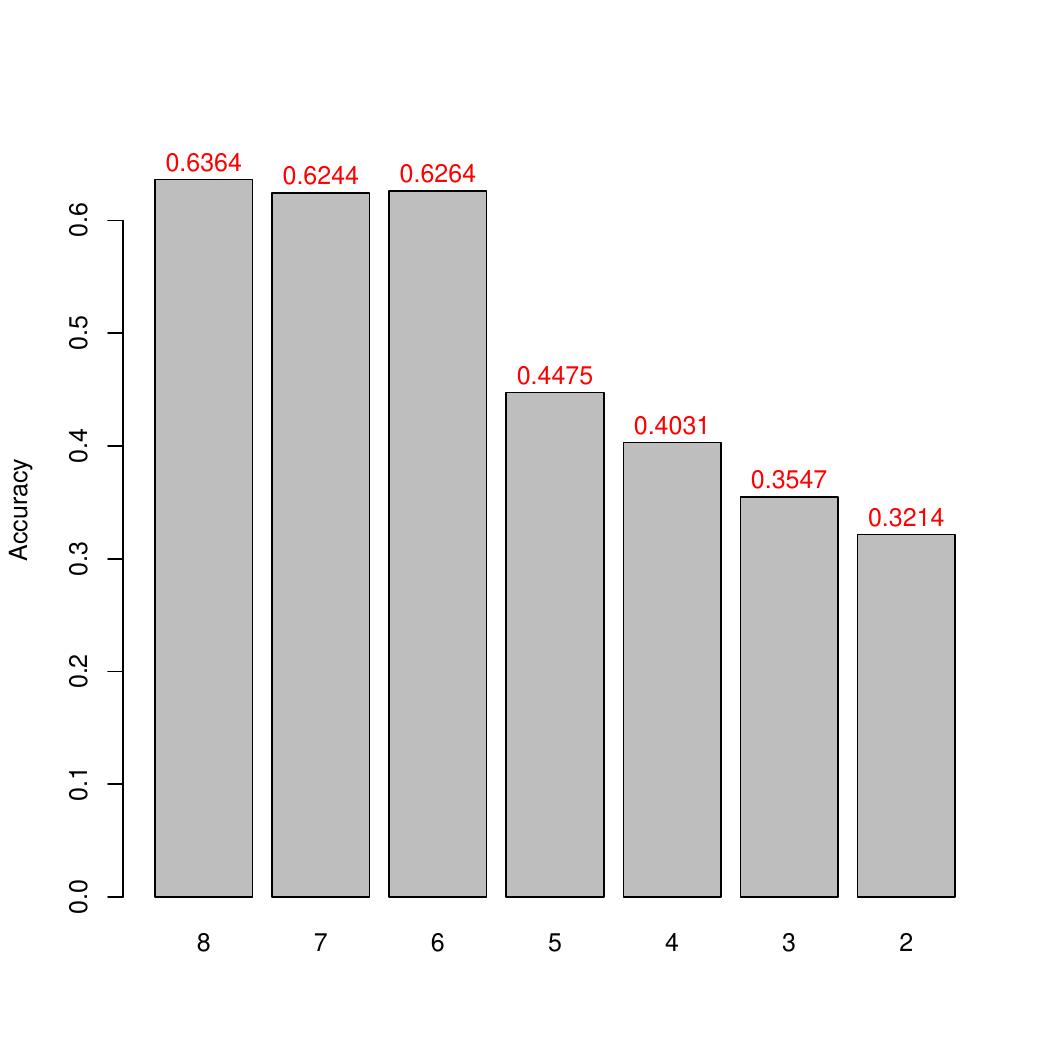}}
	\subfigure[Adjusted accuracy]{\includegraphics[width=0.48\textwidth]{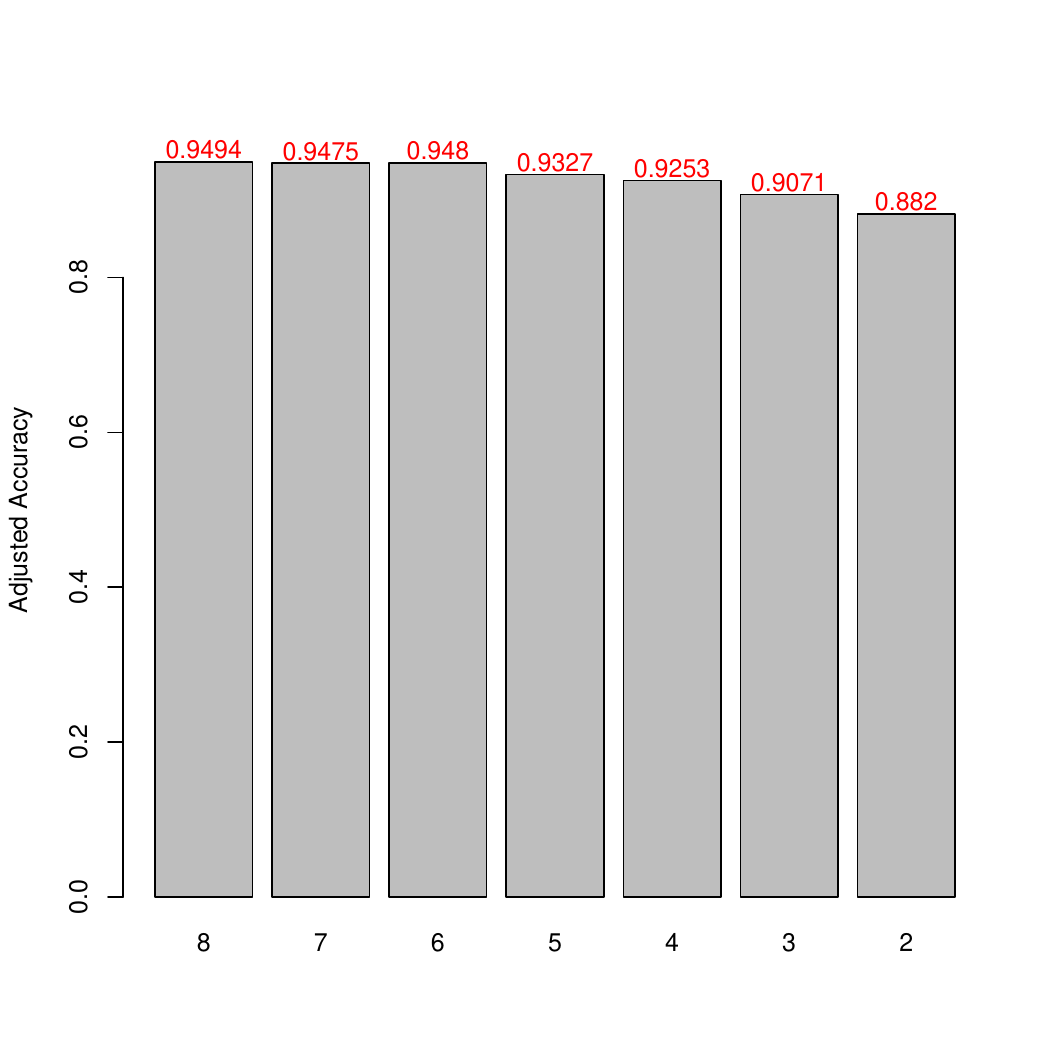}}
	\caption{The performance of the RFs classifiers.}
	\label{fig:performance}
\end{figure}

The performance of the RFs classifiers with the 7 groups of the selected variables as inputs are shown in Figure \ref{fig:performance}. It can be learned from these figures that All the performance measures of the RFs classifiers with the first three groups of variables as inputs are comparable and better than that of the classifiers with the other four groups of variables as inputs. And the performance of the other four classifiers are decreased gradually as the number of the variables are decreased.

\begin{table}
	\centering
	\caption{The selected variables according to the estimated CEs.}
	\begin{tabular}{c|l}
		\toprule
		\#group&selected variables\\
		\midrule
		8&NM\_M,Depth,PE,GR,ILD\_log10,PHIND,DeltaPHI,RELPOS\\
		7&NM\_M,Depth,PE,GR,ILD\_log10,PHIND,DeltaPHI\\
		6&NM\_M,Depth,PE,GR,ILD\_log10,PHIND\\
		5&NM\_M,Depth,PE,GR,ILD\_log10\\
		4&NM\_M,Depth,PE,GR\\
		3&NM\_M,Depth,PE\\
		2&NM\_M,Depth\\
		\bottomrule
	\end{tabular}
	\label{t:selectedvars}
\end{table}

\section{Discussion}
\label{sec:discussion}
In the experiments we use CE to measure the correlations between geological variables and facies classes. From Figure \ref{fig:pairs}, one can learn that the relationships between geological variables and facies classes are nonlinear and therefore traditional correlation measures are not applicable to this problem. CE is defined as a model-free dependence measure and also has non-parametric estimator without assumptions on the underlying distribution and therefore universally applicable. Since it is rigorous defined, it makes CEs a more reliable criteria for variable selection than traditional methods, such as cross-validation, Pearson correlation coefficient.

We select the geological variables according to their corresponding negative CE values. It can be learned from Figure \ref{fig:ce} that the negative CE of the variable ``RELPOS" is almost zero and therefore can be considered as unimportant because it carry no information about facies. The negative CE of the variable ``DeltaPHI" is below 0.2 and is also considered as unimportant because it carry less information about facies. The negative CEs of the other 6 variables are all above 0.2 and therefore are considered as important. This criteria of variable selection is validated by the prediction results that the performance of the classifiers without the variables ``RELPOS" and ``DeltaPHI" are comparable to that of the classifier with all the variables as inputs, which means that these two variables provides no information about facies and therefore is useless in classification. 

Since CE is a dependence measure with physical meanings, the result with it is interpretable. This is an advantage of CE based method compared to traditional methods based on black-box models, such as deep learning \cite{Noh2023,Dramsch2018}. In this experimental results, the largest negative CE is that corresponding to the variable ``NM\_M" which is a binary mark after examination of experts on marine and non-marine facies and therefore is reliable for classification. The second largest negative CE corresponds to the variable ``Depth", which is also reasonable since different facies classes tend to be related to specific depth ranges where a special type of sedimentary rock was deposited \cite{Pickering1986,Purkis2015}. This interpretation can be testified by Figure \ref{fig:shrimplin}, in which the distributions of facies classes along the depth of well logs are illustrated. The variable ``RELPOS", i.e., relative formation position, is confirmed to be useless for facies classification in our experiments, which is, to our knowledge, unfounded by others.

\section{Conclusions}
\label{sec:con}
In this paper we proposed a method for facies classification with CE, in which the correlations between geological variables and facies classes are measured with CE and then the variables with large negative CEs are selected for classification. We verified our method on a typical dataset for facies classification and experimental results show that our method can select less geological variables for facies classification without sacrificing performance. The geological variables such selected are also interpretable to geologists with geological meanings due to the rigorous definition of CE.

\bibliographystyle{unsrt}
\bibliography{facies}

\end{document}